\documentclass{article}

     \usepackage[final,nonatbib]{neurips_2020}

\usepackage[utf8]{inputenc} % allow utf-8 input
\usepackage[T1]{fontenc}    % use 8-bit T1 fonts
\usepackage{hyperref}       % hyperlinks
\usepackage{url}            % simple URL typesetting
\usepackage{booktabs}       % professional-quality tables
\usepackage{amsfonts}       % blackboard math symbols
\usepackage{nicefrac}       % compact symbols for 1/2, etc.
\usepackage{microtype}      % microtypography

\usepackage{xcolor}
\usepackage{adjustbox}
\usepackage{amssymb}
\usepackage{amsmath}

\newcommand{\vspacefigcaptop}{\vspace{-4mm}}
\newcommand{\vspacefigcapbot}{\vspace{-5mm}}
\newcommand{\vspacetabbot}{\vspace{-8mm}}

\title{Multi-class segmentation under severe class imbalance: A case study in roof damage assessment}

\author{%
  Jean-Baptiste~Boin\thanks{Corresponding author} \\
  CrowdAI \\
  San Francisco, CA \\
  \texttt{jb@crowdai.com} \\
  \And
  Nat~Roth \\
  CrowdAI \\
  San Francisco, CA \\
  \texttt{nat@crowdai.com} \\
  \And
  Jigar~Doshi \\
  CrowdAI \\
  San Francisco, CA \\
  \texttt{jigar@crowdai.com} \\
  \AND
  Pablo~Llueca \\
  CrowdAI \\
  San Francisco, CA \\
  \texttt{pau@crowdai.com} \\
  \And
  Nicolas~Borensztein \\
  CrowdAI \\
  San Francisco, CA \\
  \texttt{nic@crowdai.com} \\
}

\begin{document}

\maketitle

\begin{abstract}
The task of roof damage classification and segmentation from overhead imagery presents unique challenges. In this work we choose to address the challenge posed due to strong class imbalance. We propose four distinct techniques that aim at mitigating this problem. Through a new scheme that feeds the data to the network by oversampling the minority classes, and three other network architectural improvements, we manage to boost the macro-averaged F1-score of a model by 39.9 percentage points, thus achieving improved segmentation performance, especially on the minority classes.
\end{abstract}

\vspace{-1mm}
\section{Introduction and related work}
\vspace{-1mm}

As the ability to monitor and aggregate drone and satellite data has progressed, interest in leveraging these technologies to respond to natural disasters has rapidly grown. Disasters often require quick responses and these data sources are often too large or quickly-changing to be annotated by humans alone in sufficiently short time scales to assist response efforts. This challenge has inspired a wealth of work on automated semantic understanding of disaster imagery \cite{doshi2018satellite,doshi2019firenet,ben2019inundation}. 

We consider the task of building roof damage assessment \cite{gupta2019creating,xu2019building,weber2020building}. Although this task is related to semantic segmentation, since it is effectively a pixel classification task, disaster response and infrastructure detection more broadly present a number of unique difficulties that hinder the performance of classic semantic segmentation approaches. Indeed, after a natural disaster, it may be that few buildings are totally destroyed while many are unaffected. In general, objects of interest may not be present in most areas or may be far apart \cite{yu2018deepsolar}. This sparsity results in severe class imbalance. Previous works on building damage detection have noted the issues that this class imbalance and rarity pose \cite{cornebise2018witnessing,seo2019revisiting}. A number of suggestions to tackle this issue have been provided such as bagging \cite{seo2019revisiting}, pursuing a few-shot learning approach \cite{oh2019explainable} or just adopting more robust metrics \cite{xu2019building}. 

\begin{figure}
\begin{center}
\raisebox{-0.5\height}{\includegraphics[width=.21\linewidth]{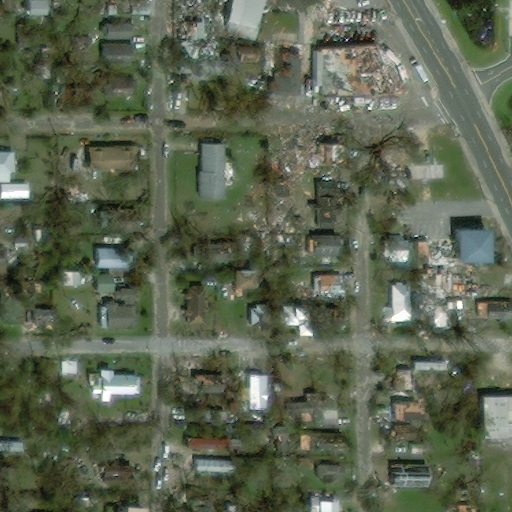}}.
\hspace{1cm}
\raisebox{-0.5\height}{{
\setlength{\fboxsep}{0pt}
\setlength{\fboxrule}{.5pt}
\fbox{\includegraphics[width=.21\linewidth]{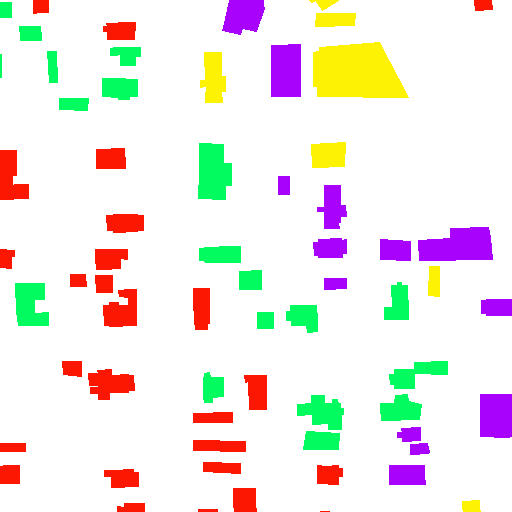}}
}}
\hspace{4mm}
\raisebox{-0.5\height}{\includegraphics[height=.13\linewidth]{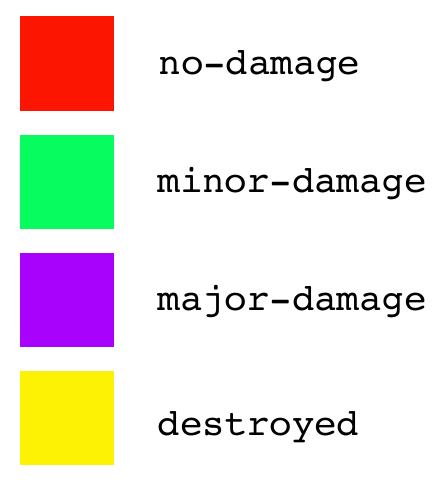}}
\end{center}
\vspace{2mm}
\vspacefigcaptop
\caption{Sample from the dataset, at the size and resolution ($512\times512$) it is fed into our models, along with the corresponding annotation, where each damage class is drawn in a different color.}
\vspacefigcapbot
\label{fig_dataset_samples}
\end{figure}

In this work, we propose a variety of techniques to combat the issues of class imbalance in the context of roof damage assessment. In particular, via ablation studies, we evaluate the effects of different neural architecture improvements and sampling methods and find an approach that significantly boosts performance on rare classes, greatly alleviating the issues described above. We demonstrate this improvement on a diverse dataset of roof damage spanning many different disasters (Fig. \ref{fig_dataset_samples}). We hope this work can inform and direct future experiments and model deployments and ultimately get faster, more effective relief to those in need.

\vspace{-1mm}
\section{Methods}
\vspace{-1mm}

Our models are based on the U-Net network architecture \cite{ronneberger2015u}. We deviate from the original architecture in a few ways: we use 9 downsampling steps instead of 4; at each resolution we use 3 convolutions (instead of 2); the convolutions at all resolutions use a fixed number of filters $N = 64$ and they are followed with the ReLU activation and a batch normalization layer; we do not include the skip connection at the full resolution stage, so the last one happens at half the resolution. Finally, all convolutional layers are padded which ensures that the input and output have the same size. We distinguish two parts of this architecture that we call the U-Net backbone (which includes all the layers up to the last skip connection) and the U-Net head (three extra convolution/activation/batch normalization blocks, each using $N$ filters, with one upsampling operation before the last block). This model extracts a feature map of size $H \times W \times N$, where $H$ and $W$ are resp. the height and width of the input patch.

The baseline model, represented in Fig. \ref{fig_network_architectures}(a), computes the class logits from this feature map by using a convolutional layer with $C+1$ filters, (where $C$ is the number of foreground damage classes, with an extra logit computed for the background class). The resulting $H \times W \times C$ feature map is then passed through a softmax layer to compute per-pixel class probabilities. For each foreground class $c$, we compute the Dice loss \cite{sudre2017generalised} between the $c$-th slice of the feature map and the binary ground truth map for class $c$. The total loss is obtained by taking the sum of the per-class losses. Note that in our formulation there is no loss computed for the background class. That slice of the activation map is implicitly used in the softmax normalization.

During training, we provide data by presenting each training sample once per epoch and shuffling them between epochs.

A more straightforward baseline would involve cross-entropy loss but preliminary experiments found that this performed much worse so we did not experiment further with this loss and did not include those results. This is actually a known problem of this loss when using very imbalanced data \cite{sudre2017generalised}. The Dice loss was popularized for binary segmentation problems to solve this specific issue, and we found that it is indeed much better suited for our task and data.

In this work we propose various improvements to considerably increase the segmentation performance of this baseline on the data. We will present each improvement independently but they can be combined as desired.

\subsection{Mitigating class imbalance with oversampling}

The first improvement relates to how we select the mini-batches used to train the model. The motivation behind this is to bypass a characteristic of the Dice coefficient/loss that occurs in binary segmentation. If a sufficient number of batches do not have a foreground sample, the average batch loss can reach a local minimum by predicting all pixels as background, which causes the model to converge to a degenerate configuration. We justify this in Appendix \ref{app_degenerate_dice}.

In our multi-class setup, this insight guides us to add a constraint that all batches must contain at least one positive example of each damage class. As long as the batch size is larger than the number of foreground classes, this can be done easily. First,
for each class $1, ..., C$, we create overlapping sets $S_1, ..., S_C$ by selecting all samples in the training set which contain pixels of the corresponding class, as well as the set $S_0$ of all the samples that only contain background pixels. We then produce batches by picking one sample of each set $S_c$ ($1 \leq c \leq C$) in every batch, and filling the rest by ensuring that all samples of $S_0$ are seen at least once. We consider that one epoch is over once all samples have been seen once. Note that because of this sampling strategy, samples containing less-common classes may be seen more than once in an epoch. We use the optimization toolbox in SciPy \cite{2020SciPy-NMeth} to find the allocation that minimizes the number of batches in an epoch. We allocate all batches at the beginning of an epoch and shuffle their order. This process does not cause noticeable overhead since it is only done once per epoch.

\subsection{Network architecture}
\label{sec_network_architecture}

Our three other improvements to the baseline model relate to the network architecture.

\begin{figure}
\makebox[\linewidth][c]{\includegraphics[width=1.3\linewidth]{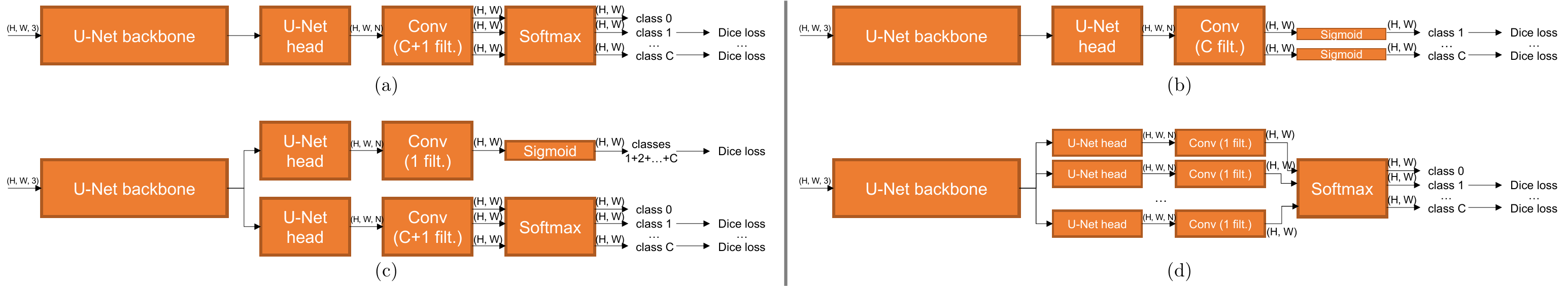}}
\vspacefigcaptop
\caption{(a) Baseline architecture. (b) Final sigmoid activation. (c) Auxiliary head. (d) Class-specific separate heads. $H$ and $W$ are resp. the height and width of the input patch, $N$ is the number of filters used in the U-Net part of the model, $C$ is the number of foreground classes.}
\vspacefigcapbot
\label{fig_network_architectures}
\end{figure}

\subsubsection{Auxiliary head}

One improvement is the addition of an auxiliary head that shares the same U-Net backbone as the rest of the model, which is trained on a class-agnostic binary roof segmentation task: the foreground class is the union of all the pixels belonging to any damage class. This extra head uses a sigmoid as the final activation and is also trained using Dice loss. In the final loss computation, this head is given a loss weight equal to $C$, in order to match the magnitude of the loss from the main head. This architecture is illustrated in Fig. \ref{fig_network_architectures}(c).

The motivation behind this is to get the rooftop segmentation task and the rooftop pixel damage classification task more decoupled. Through this auxiliary task, the model learns better features for rooftop segmentation. This, in turn, benefits the outputs corresponding to each damage class. The auxiliary head is only used during training and can be ignored for inference.

\begin{table}[b]
\caption{Statistics of the training data}
\begin{adjustbox}{center}
\small
\begin{tabular}{lcccc}
\toprule
Class name & \# non-empty patches & Ratio of non-empty patches & Pixel distribution & Foreground pixel distribution \\
\cmidrule(lr){1-1}
\cmidrule(lr){2-5}
\texttt{background} & 34619 & 100.0\% & 96.36\% & - \\
\texttt{no-damage} & 12599 & 36.4\% & 2.78\% & 76.5\% \\
\texttt{minor-damage} & 4259 & 12.3\% & 0.33\% & 9.0\% \\
\texttt{major-damage} & 4129 & 11.9\% & 0.34\% & 9.3\% \\
\texttt{destroyed} & 3881 & 11.2\% & 0.19\% & 5.2\% \\
\bottomrule
\end{tabular}
\end{adjustbox}
\label{tab_dataset_stats}
\end{table}

\subsubsection{Class-specific separate heads}

The next proposed improvement is to replace the single multiclass head with $C+1$ binary heads that split off from each other much earlier in the architecture: the split happens right after the U-Net backbone, as is shown in Fig. \ref{fig_network_architectures}(d). Head $c$ produces a $(H, W)$ feature map corresponding to class $c$, and the probability distributions per pixel can be obtained as usual by concatenating those feature maps and computing the softmax across the class dimension.

Our hypothesis is that by doing so the model can learn more complex class-specific filters in each head. Since we do not reduce the number of filters in the U-Net heads, we note that this effectively increases the capacity of the model.

\subsubsection{Final sigmoid activation}

Our last proposed modification is to replace the final softmax activation with a sigmoid activation, as illustrated in Fig. \ref{fig_network_architectures}(b). This change makes the classes more independent from each other which prevents cross-talk between classes. We note that this change also makes the background logit unnecessary since it is not used in any loss, so we can just drop that output without adverse effects, making the model lighter.

Changing the activation makes the model more similar to a multi-label segmentation model, since now a given pixel could get a high score for multiple classes at once. So, at inference time we need to add an extra step to merge the class-specific binary predictions into a single class label. We assign a pixel to foreground or background based on its maximum post-sigmoid score across all foreground classes. If that score is larger than a threshold $t_{fg}$ (set as $0.2$ in our experiments), the prediction for that pixel is the class that maximizes the score; otherwise it is the background class.

\vspace{-1mm}
\section{Experiments and results}
\vspace{-1mm}

\subsection{Experimental setup}

\begin{table}
\caption{Performance of the model using various setup combinations. The best performance values are shown in bold.}
\begin{adjustbox}{center}
\small
\begin{tabular}{lcccccc}
\toprule
Model \# & Oversampling & Aux. head & Separate heads & Sigmoid & micro F1-score (\%) & macro F1-score (\%)\\
\cmidrule(lr){1-1}
\cmidrule(lr){2-5}
\cmidrule(lr){6-7}
0 (baseline) &  &  &  &  & 60.2 & 17.4 \\
1 &  & \checkmark &  &  & 59.8 & 17.1 \\
2 &  &  &  & \checkmark & 59.8 & 32.4 \\
3 &  & \checkmark & \checkmark & \checkmark & 61.9 & 37.6 \\
4 &  &  & \checkmark & \checkmark & 65.8 & 52.1 \\
5 & \checkmark &  &  & \checkmark & 63.6 & 52.4 \\
6 & \checkmark &  &  &  & 65.2 & 53.3 \\
7 & \checkmark & \checkmark &  & \checkmark & 66.6 & 54.5 \\
8 & \checkmark & \checkmark &  &  & 67.1 & 54.8 \\
9 & \checkmark &  & \checkmark & \checkmark & 67.2 & 55.7 \\
10 (ours) & \checkmark & \checkmark & \checkmark & \checkmark & \textbf{69.5} & \textbf{57.3} \\
\bottomrule
\vspacetabbot
\end{tabular}
\end{adjustbox}
\label{tab_results}
\end{table}

We use a subset of the xBD dataset \cite{gupta2019creating} for our experiments. This dataset contains satellite imagery from 19 natural disasters where building rooftops are annotated with four levels of damage ($C = 4$): \texttt{no-damage}, \texttt{minor-damage}, \texttt{major-damage} and \texttt{destroyed}. For this work we only used the post-disaster imagery. We used $8466$ images for our training set and $1504$ for our validation set, each split into non-overlapping $512 \times 512$ patches. Some statistics of our training dataset are reported in Table \ref{tab_dataset_stats} and highlights two levels of imbalance that we face. First, the foreground-background imbalance can be seen from the pixel distribution: $96.4\%$ of the pixels are background pixels. Second, a large majority of the foreground pixels ($76.5\%$) belong to the \texttt{no-damage} class.

We use the same training setup for all the models we show results for in this work. We train on a single GPU using a batch size of 8 and the Adam optimizer \cite{kingma2015adam}, as well as random rotations and horizontal-vertical flips for data augmentation. When performing oversampling, an epoch has more iterations than when not oversampling, so we redefine our training epochs as being a fixed $1500$ iterations, so that the training regime is comparable for all conditions. We train the models for $200$ training epochs, computing the validation loss across the validation set after each one. We use a step decay learning rate schedule, with an initial learning rate of $2\mathrm{e}{-4}$ that we divide by $2$ every $50$ training epochs.

The F1-score is typically the metric of choice for binary segmentation. In the multi-class case, it can be extended to two different scores that exhibit different properties: the micro-averaged F1-score, which is biased by class frequency and is thus better suited for evaluating the overall performance of the model, and the macro-averaged F1-score, which is the average of the F1-scores obtained for each foreground class and is thus better at evaluating how well the model handles imbalance. For each model we pick the model weights that minimize the validation loss and we report both F1-scores.

\subsection{Ablation study}

In order to evaluate the impact of each one of our proposed improvements, we combine them in 11 different configurations and train a model for each configuration. Results are reported in Table \ref{tab_results}, with the baseline shown at the top (model \#0) and our final model with all the improvements shown at the bottom (model \#10), the other ones being ordered by increasing macro-averaged F1-score.

Satisfyingly, the best scores are obtained for model \#10, when all four improvements are added, which shows that they combine well. The micro-averaged F1-score improves by 9.3 percentage points (p.p.) compared to the baseline. The most striking gain is on the macro-averaged F1-score, which improves by 39.9 p.p. The reason that the this value is so low for the baseline (and model \#1 as well) is that the F1-score for the minority classes (\texttt{minor-damage}, \texttt{major-damage}, \texttt{destroyed}) is actually zero, and only the F1-score for the majority class is non-zero. Because of class imbalance, this model completely fails at predicting these classes.

It is important to note that due to a limited computational budget we only trained one model per configuration, and therefore some of the counter-intuitive ordering between configurations could be explained by being (un)lucky in the random initialization. The most puzzling result is that model \#3 has a noticeably worse macro-averaged F1-score compared to model \#4, though we would expect it to be better since it only differs with the inclusion of the auxiliary head. Nevertheless, the results from the table paint a clear picture that the scores generally increase as more techniques are included.

Out of the techniques we present, the one that has the most impact on the final performance seems to be our improved sampling scheme. On its own (model \#6), it already improves the macro-averaged F1-score by 35.9 p.p. from the baseline. This is great news because this technique is among the least constraining out of the ones we propose, as it does not incur any additional memory cost. The other three techniques also show some noticeable gains at several occasions, as can be illustrating by comparing model pairs (\#5, \#7), (\#6, \#8) and (\#9, \#10) for the auxiliary head, model pairs (\#2, \#4) and (\#7, \#10) for the class-specific separate heads, and the pairs (\#0, \#2) for the sigmoid activation.

We show some additional results in Appendix \ref{app_aux_head} that highlight the even stronger benefit of the auxiliary head when used for binary segmentation.

\subsection{Performance of final model}

\begin{figure}
\begin{center}
\includegraphics[width=0.35\linewidth]{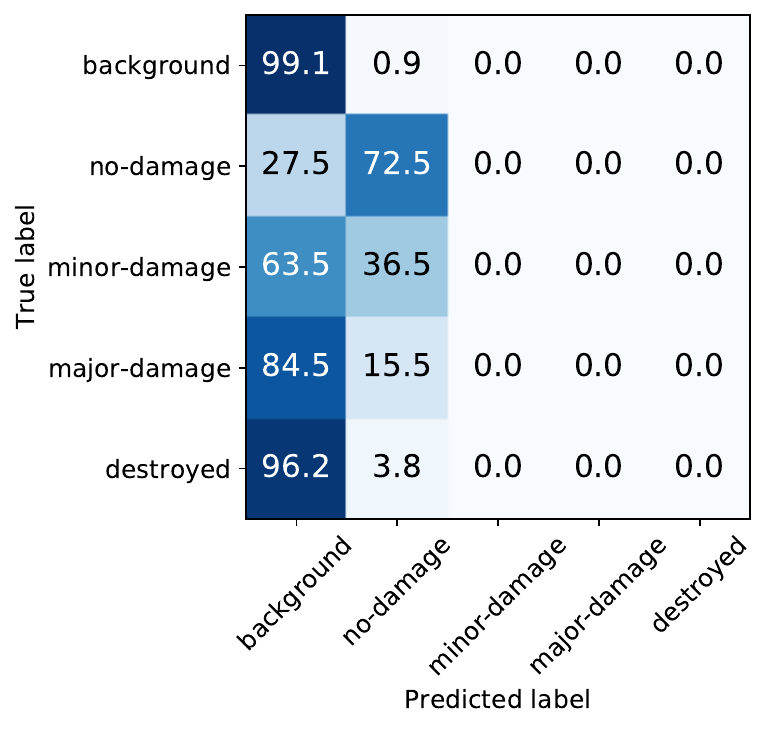}
\hspace{1cm}
\includegraphics[width=0.35\linewidth]{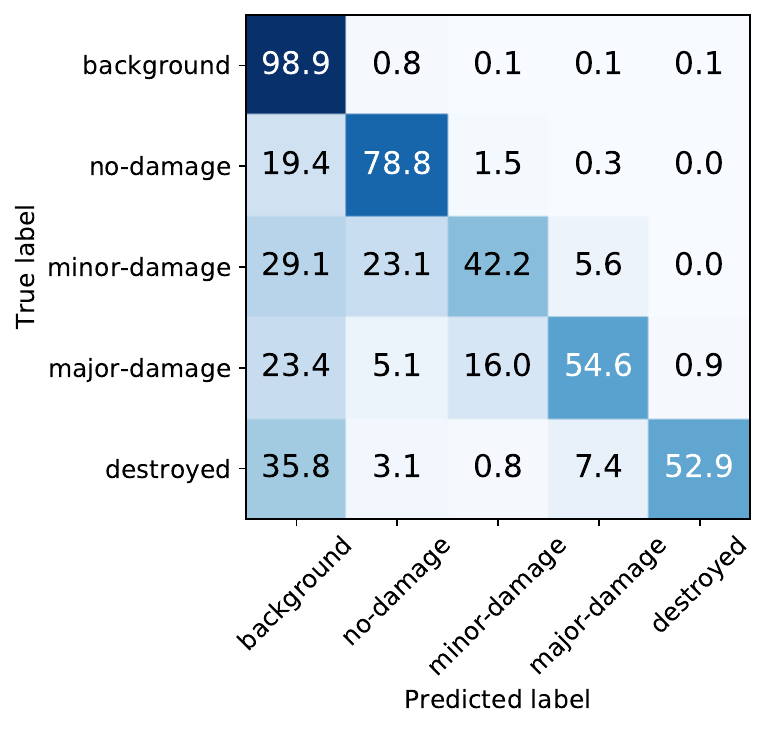}
\end{center}
\vspacefigcaptop
\caption{Normalized confusion matrices (in \%) on the validation set for the baseline model (left) and our model with all proposed improvements (right). Each row corresponds to all the pixels of a class and for each row the values represent the distribution of the predicted class of those pixels (each row sums up to 100\%). A perfect predictor would have all its diagonal elements as 100\% and off-diagonal elements as 0\%.}
\vspacefigcapbot
\label{fig_confusion_matrices}
\end{figure}

In Fig. \ref{fig_confusion_matrices} we show the confusion matrices for the baseline model (\#0) and our final improved model (\#11). This more detailed presentation of the model performance makes it clear that the baseline fails at predicting all foreground classes but the majority \texttt{no-damage} class, as we reported before.

The highest confusion of our final model seems to come from a large amount of false negatives for the \texttt{destroyed} class. This is unsurprising because some destroyed buildings may have completely disappeared from the post-disaster imagery. Combining this input with the pre-disaster imagery would help to mitigate that.

%One interesting phenomenon we observe on the confusion matrix of our model is that for the rooftop pixels correctly predicted as rooftop, there is an asymmetry for the off-diagonal elements corresponding to the two intermediate damage classes (\texttt{minor-damage} and \texttt{major-damage}). For these classes, the model tends to more often under-estimate the damage than to over-estimate it. Our hypothesis is that this is because the roofs are annotated based on damage that can be very localized. It is possible that a pixel belonging to a \texttt{minor-damage} roof is actually relatively far from the actual location of the damage. Given the convolutional nature of our networks, where the direct neighborhood of a pixel typically have a stronger impact on the predictions than more remote pixels, this explains why such a pixel could end up classified as \texttt{no-damage}.

\vspace{-1mm}
\section{Conclusion}
\vspace{-1mm}

In this work we proposed four techniques to improve multi-class segmentation in the context of roof damage assessment, where the foreground data can be sparse and the classes very imbalanced. We showed that these four techniques can be combined to significantly improve the segmentation performance, especially on minority classes. An interesting avenue for future work would be to explore whether some of these techniques could improve segmentation algorithms that use a very different paradigm such as instance segmentation.

\begin{ack}
We acknowledge the Maxar/DigitalGlobe Open Data program for the imagery used in the xBD dataset.
\end{ack}

\bibliographystyle{IEEEbib}
\bibliography{neurips_2020}

\pagebreak

\appendix
\section{Degenerate models when using Dice loss with empty batches}
\label{app_degenerate_dice}

We consider a model that performs binary segmentation. The predicted values are noted $\hat{y}_i \in [0, 1]$ while the ground truth values are noted $y_i \in \{0, 1\}$. The Dice coefficient is defined as:
\begin{align*}
DC = \frac{2 \sum_i y_i \hat{y}_i + \epsilon}{\sum_i (y_i + \hat{y}_i) + \epsilon}
\end{align*}
where $\epsilon$ is a value used to avoid the numerical issue of division by zero. The Dice loss is obtained from the Dice coefficient with: $DL = 1-DC$.

When a batch doesn't contain any foreground pixel (i.e. $y_i = 0$ for all $i$), then the Dice coefficient is:
\begin{align*}
DC = \frac{\epsilon}{\sum_i \hat{y}_i + \epsilon}
\end{align*}
A perfect Dice coefficient will only be achieved by predicting zero everywhere, and any small deviation from this will severely decrease its value, which means a huge penalty in terms of Dice loss. On the other hand, if a batch does contain foreground pixels, this degenerate model gets a loss that is bounded by 1 (by definition).

Now, if foreground pixels are only present in one batch every $k$ batches, a degenerate model that predicts zero everywhere will achieve an average loss bounded by:
\begin{align*}
DL_{\textrm{average}} &= \frac{1}{k} DL(\textrm{foreground}) + \frac{k-1}{k} DL(\textrm{no foreground}) \leq \frac{1}{k} 
\end{align*}

On the same dataset, any model with imperfect precision will most likely predict false positives on more than one of the $k-1$ batches with no foreground pixels, thus incurring a loss $\approx 1$ for these batches, meaning that the average loss will be $DL_{\textrm{average}}> 1/k$.

It then becomes apparent that the easiest way for our training process to get a better loss would be by forcing this model to become a degenerate model that only predict zeros.

\section{Benefits of the auxiliary head for binary segmentation}
\label{app_aux_head}

\begin{table}[b]
\caption{Comparison of the performance of a single-class segmentation model with or without an auxiliary head}
\begin{adjustbox}{center}
\small
\begin{tabular}{lcc}
\toprule
Class name & \multicolumn{2}{c}{F1-score (\%)} \\
\cmidrule(lr){2-3}
 & without aux. head & with aux. head \\
\cmidrule(lr){1-1}
\cmidrule(lr){2-3}
\texttt{no-damage} & 74.1 & \textbf{74.6} \\
\texttt{minor-damage} & 32.5 & \textbf{37.8} \\
\texttt{major-damage} & 46.1 & \textbf{57.2} \\
\texttt{destroyed} & 47.4 & \textbf{55.2} \\
\bottomrule
\end{tabular}
\end{adjustbox}
\label{tab_single_class_results}
\end{table}

We found that using an auxiliary head works relatively well for multi-class segmentation, but it provides even more gains when one only needs to train a single-class binary segmentation model.

For each damage class, we trained a pair of single-class networks using a similar set-up as before (U-net with sigmoid output). The first network had a single segmentation head trained on the objects from that damage class, whereas the other one had an extra auxiliary head trained on rooftops from all classes. That extra head was then discarded at inference time.

We report the F1-score obtained for each class and each pair of networks in Table \ref{tab_single_class_results}. In all cases we notice a boost in performance when using the auxiliary head. Although this boost is almost negligible for the majority class, it is significant for the minority classes. Our interpretation is that the model learns from using data from other classes. Indeed, learning good features that distinguish rooftops of any class from the background is useful for all damage classes.

\end{document}